\pdfoutput=1

\documentclass[11pt]{article}

\usepackage{ACL2023}

\usepackage{times}
\usepackage{latexsym}
\usepackage{graphicx} 
\usepackage{amsmath,amsfonts,bm}
\usepackage{enumitem}

\usepackage[T1]{fontenc}

\usepackage[utf8]{inputenc}

\usepackage{microtype}

\usepackage{inconsolata}
\usepackage{amsmath}
\usepackage{amssymb}
\usepackage{amsfonts}

\newcommand{\Minseok}[1]{}
\newcommand{\e}[1]{\mathbb{E}_{#1}}

\title{Aligning Large Language Models via Fine-grained Supervision}

\newcommand*\samethanks[1][\value{footnote}]{\footnotemark[#1]}

\author{\bf
Dehong Xu$^{1}$\thanks{\quad{}Corresponding authors. }   \thanks{\quad{}Author performed the work while interned at Amazon.},
Liang Qiu$^2$\samethanks[1],
Minseok Kim$^2$,
Faisal Ladhak$^2$,
Jaeyoung Do$^3$\\ \\
\normalfont{$^1$Department of Statistics, UCLA\quad{}} $^2$Amazon \\
$^3$Department of Electrical and Computer Engineering, Seoul National University \\ 
 \small{
   \textbf{Correspondence:} \href{mailto:xudehong1996@ucla.edu}{xudehong1996@ucla.edu}, \href{mailto:liangqxx@amazon.com}{liangqxx@amazon.com} 
 }
}

\begin{document}
\maketitle
\begin{abstract}
Pre-trained large-scale language models (LLMs) excel at producing coherent articles, yet their outputs may be untruthful, toxic, or fail to align with user expectations. Current approaches focus on using reinforcement learning with human feedback (RLHF) to improve model alignment, which works by transforming coarse human preferences of LLM outputs into a feedback signal that guides the model learning process. However, because this approach operates on sequence-level feedback, it lacks the precision to identify the exact parts of the output affecting user preferences. To address this gap, we propose a method to enhance LLM alignment through fine-grained token-level supervision. Specifically, we ask annotators to minimally edit less preferred responses within the standard reward modeling dataset to make them more favorable, ensuring changes are made only where necessary while retaining most of the original content. The refined dataset is used to train a token-level reward model, which is then used for training our fine-grained Proximal Policy Optimization (PPO) model. Our experiment results demonstrate that this approach can achieve up to an absolute improvement of $5.1\%$ in LLM performance, in terms of win rate against the reference model, compared with the traditional PPO model. 





\end{abstract}

\section{Introduction}

One key objective in advancing large language models\,(LLMs) is to ensure safe, beneficial human interaction. However, current pre-trained models, mostly trained on web and book texts, often generate biased or toxic text, misaligning with human intentions. To address this issue, numerous studies\,\cite{ouyang2022training,rafailov2023direct,bai2022constitutional,bai2022training,yuan2023rrhf,touvron2023llama,ramamurthy2022reinforcement} have integrated human feedback into the training process. A significant advancement is reinforcement learning from human feedback\,(RLHF)~\cite{ouyang2022training}, which usually consists of two phases: First, a reward model (RM) is trained from preference data, which comprises various responses alongside their human-assigned preference scores for a given prompt. Then, this reward model is applied to optimize a final model using Proximal Policy Optimization\,(PPO)~\citep{schulman2017proximal}.

Recent works\,\cite{wu2023fine,rafailov2023direct,fernandes2023bridging,guo2023beyond,wang2024secrets} discovered limitations of the current RM, specifically their misalignment with human values. This misalignment stems from two main issues: $(i)$ the presence of incorrect and ambiguous preference pairs in the human-labeled datasets; $(ii)$ the limited insight inherent in sequence-level feedback. Specifically, from a data collection standpoint, the task of comparing the overall quality of model outputs is challenging for human annotators when outputs exhibit both desired and undesired behaviors in different parts. Moreover from the RM perspective, the reliance on preference-based data labeling leads to sparse training signals. This sparsity discourages the model's ability to distinguish finer details between responses and further limits the capacity for reward optimization.

\begin{figure*}[ht]
\centering
\includegraphics[width=.99\linewidth]{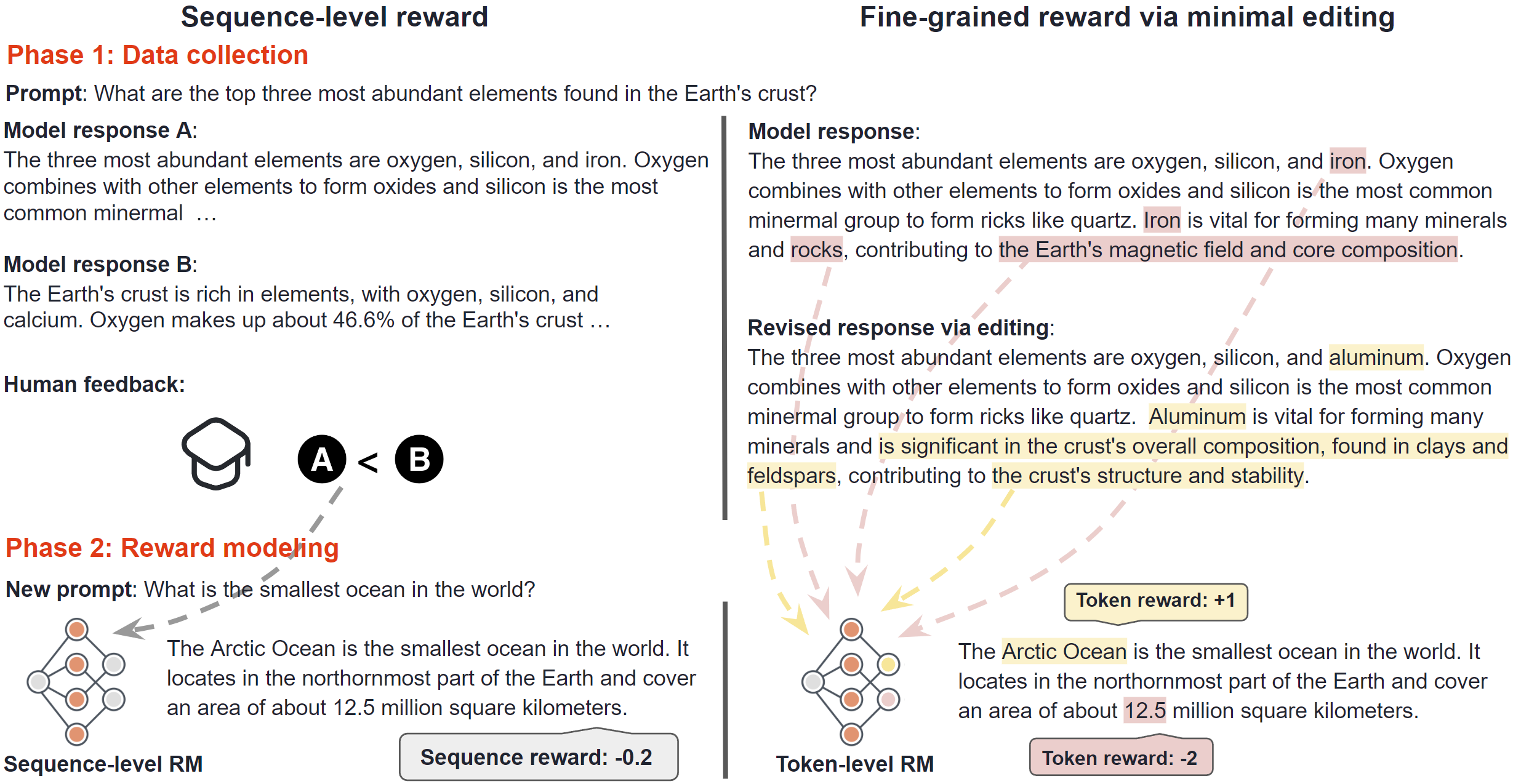} \\
  \caption{\small The comparison between sequence-level reward modeling (Left) and our method of fine-grained reward modeling via minimal editing (Right). Our approach diverges from sequence-level reward modeling in two key aspects: (1) Data Collection, where we ask a human or LLM to edit the model response; and (2) Reward Modeling, which enables our model to assign rewards to individual tokens, as opposed to assessing the entire sequence collectively. 
  }
  \label{fig: rlhf}
\end{figure*}


To tackle this challenge, we propose the following two-fold contributions as illustrated in Figure\,\ref{fig: rlhf}: 
\begin{itemize}[leftmargin=9pt, noitemsep]
\item We introduce a new data collection approach that asks annotators to edit responses from existing RM datasets to be more preferable. By comparing the original and edited responses, we obtain detailed token-level insights that are essential for training our fine-tuned reward model.
\item We propose a new token-level reward modeling approach that provides reward signals at the token level. Different from coarse-grained sequence-level rewards, our approach offers more granular feedback, pinpointing the specific parts of a response that are effective or need improvement, which hence helps RL optimization. 

\end{itemize}
Experiment results using AlpacaFarm\,\cite{dubois2023alpacafarm} environment indicate that our proposed approach improves LLMs' performance up to $5.1\%$ against the baseline in terms of win rate, given the same amount of data for training.

\section{Method}
\label{sec:method}

In this section, we introduce our approach to fine-grained data collection through editing and token-level reward modeling. 




\subsection{Fine-grained data collection via minimal editing}

The conventional RLHF pipeline, as outlined in prior works\,\cite{ouyang2022training,dubois2023alpacafarm}, involves three key stages: supervised fine-tuning (SFT), reward modeling (RM), and proximal policy optimization (PPO). In the RM phase, the standard practice entails collecting a dataset of human evaluations comparing two or more model outputs in response to a series of prompts. The dataset is represented as $\mathcal{D}=\{x^{(i)}, y^{(i)}_w, y^{(i)}_l\}^N_{i=1}$, where $x$ denotes a prompt and $(y_w, y_l)$ indicates the preferred and less preferred responses, respectively. Utilizing such a dataset, earlier RLHF research focused on developing a reward model $R_\phi$ that determines the more favored model output. This holistic reward model associates each input prompt $x$ and its corresponding output $y$ with one scalar value reflecting the output's overall quality.

 
However, as shown in the left panel of Figure~\ref{fig: rlhf}, annotating a pair of model outputs that are substantially different can be a difficult task for humans, especially when each response exhibits a mix of desirable and undesirable behaviors. To address this issue, we introduce a novel data collection technique aimed at obtaining fine-grained supervision, which offers richer, comparative information beyond simple binary choices. Instead of annotating entire responses, our method involves targeted editing by humans or language models, as depicted in the right panel of Figure~\ref{fig: rlhf}. The goal is to retain the majority of the original response while making improvements to specific areas in need of enhancement. Specifically, we introduce a response editing process in which we ask humans or prompt LLMs to perform targeted modifications. For fine-grained data collection, our method works for both human annotators and language models, following~\cite{ding2022gpt,gilardi2023chatgpt,wang2022self,chiang2023can}. 

In practice, we prompt a proprietary LLM, such as \textit{Claude-2}~\cite{bai2022constitutional}, to apply edits to the original output. In the experiment, the original preference pairs $(y_w, y_l)$ were not included and we only utilized $y_l$ from the original dataset for minimal editing. This approach maintains the same amount of data as the baseline methods, ensuring a fair comparison. Details of the prompt used for editing can be found in Appendix~\ref{sec:appendix1}, and the examples of fine-grained annotation with minimal editing are shown in Appendix~\ref{sec:appendix2}. Our method is based on the assumption that the edits inherently improve a response, making changes only when they enhance alignment with human values. The approach enables the refinement of responses by providing clear insights into the specific areas that require improvement. 



\subsection{Token-level reward modeling}
In this section, we will first introduce the RL environment and then define our token-level reward modeling scheme. 

Language generation can be defined as a Markov Decision Process (MDP) $\langle\mathcal{S}, \mathcal{A},\mathcal{R},\mathcal{P}, \mathcal{\gamma}\rangle$. $\mathcal{S}$ refers to the state space and we define the start state $s_1$ as the input prompts $\{x\}$. An action at t-step $a_t$ is a generated token. The transition function of the environment is denoted as $\mathcal{P}: \mathcal{S}\times\mathcal{A}\rightarrow\mathcal{S}$, $s_t=\{x, a_1, ..., a_{t-1}\}$. A response $y$ of length $T$ is then $y=\{a_1, ..,a_{T}\}$. In our token-level reward scheme, a reward is assigned to each generated token $a_t$ by   $\mathcal{R}\colon\mathcal{S}\times\mathcal{A}\rightarrow\mathbb{R}$, where at each time step $t$ there is a learned reward function $r_t=r_\phi(s_t, a_t)$. Therefore, for each response, we have a trajectory $\tau=\{s_1,a_1,r_1,...s_t,a_t,r_t, ...s_{T},a_T, r_T\}$. 

We define the reward of the whole trajectory as the average of rewards assigned to each token:
\begin{align}
\begin{split}\label{eq:prob}
R(\tau)
&=\frac{1}{T}\sum_{t=1}^T r_t. 
\end{split}
\end{align}
Following the Bradley-Terry (BT) model\,\cite{bradley1952rank} for preference modeling, we formulate the distribution of human preference for responses as below: 
\begin{align}
\begin{split}
  p(\tau^i \succ \tau^j)&=\frac{\exp(R(\tau^i))}{\exp(R(\tau^i)) + \exp(R(\tau^j))}\\
  &=\sigma(R(\tau^i)-R(\tau^j)), 
\end{split}
\end{align}
where $\tau^i$ and $\tau^j$ represent two different responses generated from the same prompt. Under the setting of our fine-grained supervision dataset, we assume $\tau^i$ only makes edits on $\tau^j$ while maintaining most parts unchanged. We define $U_0=\{t|a_t^i = a_t^j\}$ and $U_1=\{t|a_t^i\neq a_t^j\}$ to represent the unchanged and changed parts. 

Regarding the reward model as a binary classifier, we use negative log-likelihood as the loss function. By plugging in Equation~\ref{eq:prob}, we have: 
\begin{align}
\begin{split}
    &\mathcal{L}=-\e{(\tau^i,\tau^j)\sim\mathcal{D}} \left[\log\sigma(R(\tau^i)-R(\tau^j))\right]\\
    &=-\e{(\tau^i,\tau^j)\sim\mathcal{D}} [\log\sigma((\frac{1}{T^i}-\frac{1}{T^j})\sum_{t\in U_0} r_t \\ 
    &+ \frac{1}{T^i}\sum_{t\in U_1} r_t^i - \frac{1}{T^j}\sum_{t\in U_1} r_t^j)],
\end{split}
\end{align}

Ideally, we aim for the unchanged part to maintain a consistent reward. Under this assumption, and if the two responses are of equal length, the first term of the loss function can be removed:

\begin{align}
\begin{split}\label{eq:approx}
    \mathcal{L}\approx-\e{(\tau^i,\tau^j)\sim\mathcal{D}}[\log\sigma(\frac{1}{T^i}\sum_{t\in U_1} r_t^i - \frac{1}{T^j}\sum_{t\in U_1} r_t^j)]
\end{split}
\end{align}
For the edited part, the loss function is thus designed to maximize the reward for the preferred response and minimize it for the less favored one.

With a trained token-level reward model, we can integrate it into the Proximal Policy Optimization (PPO)~\citep{schulman2017proximal} algorithm. In the traditional PPO-RLHF method, each token in the sequence is assigned a reward of the form $[-KL_1, -KL_2, ..., R-KL_n]$, where $KL_i$ denotes the Kullback-Leibler divergence~\cite{kullback1951information} for the generated token sequence up to that point, and $R$ represents the sequence-level reward from the reward model. Generalized Advantage Estimation (GAE)~\citep{schulman2015high} is then employed to calculate the advantage at the token level. 

In contrast, our approach assigns a reward $R_i$ directly from the token-level reward model to each token in the sequence, resulting in a reward vector of $[R_1, R_2, ..., R_n]$. This approach enhances the granularity of feedback at each step of the sequence generation process, without changing the underlying GAE and policy update procedure. Consequently, the computational cost remains comparable to the standard RLHF approach.

\section{Experiments}

In this section, we demonstrate our experimental setup and empirical results in detail. 

\subsection{Experimental setup}

In constructing our dataset, we follow the framework established by AlpacaFarm\,\cite{dubois2023alpacafarm}, which offers a simulation environment that includes data splits for SFT, RM, PPO, and evaluation processes. Building on this, we develop our refined RM dataset using the fine-grained approach, where we employ \textit{Claude-2} ~\cite{bai2022constitutional} to perform targeted editing. Edits are generated on the less preferred responses from the original pairwise data, ensuring lightweight yet effective modifications.

We evaluate our method by finetuning the pre-trained \textit{LLaMA-7B}\,\cite{touvron2023llama} model. To assess the quality of our model's generation compared to baseline models, we employ a win-rate measurement, where the model $p_\theta$ is evaluated against a reference model $p_{\text{ref}}$. This method involves pairwise comparisons to estimate how often $p_\theta$'s outputs are preferred over $p_{\text{ref}}$'s for given instructions. Both our model and the baselines are evaluated against the same reference model, \textit{Davinci003}, aligning with AlpacaFarm\,\cite{dubois2023alpacafarm}. To assess the win rate, we employ \textit{Claude} as the judge, following the simulated approach in~\cite{zheng2023judging}. 

\begin{table}
\centering
\begin{tabular}{ll}
\hline
Model & Win rate (\%) \\
\hline
Fine-grained Token-level PPO & $\pmb{51.6 \pm 1.8}$ \\
Fine-grained PPO & $\pmb{51.2 \pm 1.8}$ \\
Davinci003~\cite{brown2020language} & $50.0$ \\
PPO-RLHF~\cite{ouyang2022training}  & $46.5 \pm 1.8$\\
\hline
\end{tabular}
\caption{Evaluation results by \textit{Claude}. \textit{Davinci003} is the reference model. All results of other models are from~\cite{dubois2023alpacafarm}. }\label{table:eval}
\end{table}



To evaluate the effectiveness of our data annotation approach and token-level reward model, we train two models: $(i)$ \textbf{Fine-grained PPO} that only uses our fine-grained RM dataset with editing while still trained with a sequence-level reward, and $(ii)$ \textbf{Fine-grained Token-level PPO} that incorporates both the fine-grained RM dataset and token-level reward modeling, and hence applies token-level reward to PPO. 

\subsection{Experiment results}
\paragraph{Results in human value alignment}


Table~\ref{table:eval} showcases our methods (highlighted) alongside the baseline PPO-RLHF model, both trained on \textit{LLaMA-7B}~\cite{touvron2023llama}. Results indicate that our novel data collection technique, when integrated with standard PPO training, leads to an absolute performance increase of $4.7\%$ compared to traditional methods (refer to lines 2 vs. 4). This highlights the effectiveness of our fine-grained data collection strategy. Moreover, when trained with the same fine-grained dataset, the token-level reward model (line 1) demonstrates further alignment improvements compared to the PPO alone (line 2), indicating the importance of token-level rewards. Together, these findings affirm that our approach significantly outperforms the traditional PPO-RLHF model.



\paragraph{Reward model analysis}

\begin{table}
\centering
\begin{tabular}{lc}
\hline
Model & Accuracy (\%) \\
\hline
RM w/ Fine-grained dataset & $\pmb{85.2 \pm 1.8}$ \\
RM w/o Fine-grained dataset  & $58.2 \pm 1.8$\\
\hline
\end{tabular}
\caption{Reward model accuracy. Leveraging the fine-grained dataset enhances the reward model's ability to assign correct rewards to responses. }\label{table:reward}
\end{table}

To explain the observed performance increase, we further investigate the effectiveness of the reward model. We test its accuracy in assigning higher rewards to superior responses within the evaluation set. As shown in Table~\ref{table:reward}, our fine-grained dataset enables the learned reward model to reach an accuracy of approximately $85.2\%$, outperforming the model trained with the original dataset. This result demonstrates that our data collection method enhances the capability of our reward model to identify and appropriately reward better responses.

\paragraph{Training efficiency}

\begin{table}
\centering
\begin{tabular}{llc}
\hline
Model & Step & Tr. hours\\
\hline
RLHF~\cite{ouyang2022training} & RM & $0.2$  \\
Fine-grained RLHF & RM & $\pmb{0.3}$  \\
\hline
RLHF~\cite{ouyang2022training} & PPO & $4$  \\
Fine-grained RLHF & PPO & $\pmb{2}$  \\
\hline
\end{tabular}
\caption{Training efficiency. Highlighted numbers represent the training hours (Tr. hours) of the fine-grained PPO model trained with token-level rewards. }\label{table:training}
\end{table}

Table~\ref{table:training} illustrates the training costs for different models. Note that all the models are trained on 8 NVIDIA A100 GPUs (80G) with the same batch size for both phases. While the training time for the reward modeling phase is comparable between our method and the baseline, our fine-grained reward model significantly boosts the efficiency of RL optimization. It reduces the time required for PPO to converge to its optimal performance by half, due to our more precise and fine-grained reward function. 
Based on the experiment results, our reward function can provide more accurate and denser training signals, which can help RL algorithms converge faster. This improvement in training efficiency could be important for LLM alignment, especially when the size of the LLM becomes increasingly large. 





\section{Limitations}

Although the empirical results show that our approach achieves better performance in model alignment, we struggle to provide rigorous mathematical proof to conclusively demonstrate the effectiveness of this reward allocation strategy, specifically in Equation~\ref{eq:approx}. 

\section{Conclusion}
In this paper, we introduce a fine-grained RLHF framework that includes a data collection technique alongside a token-level reward model. This approach enables better value alignment by learning a more accurate reward model, facilitating faster convergence for PPO. Our experimental results show performance improvement based on automatic evaluations compared to the baseline method. 

\section*{Acknowledgments}

We would like to thank Yi Xu, Puyang Xu and other members of Amazon, as well as Ying Nian Wu and Minglu Zhao and from University of California, Los Angeles for their valuable discussions and constructive feedback. Dehong Xu’s research for this work was financially supported by Amazon during his internship at Amazon.



\newpage






\bibliography{anthology,custom}

\begin{thebibliography}{22}
\expandafter\ifx\csname natexlab\endcsname\relax\def\natexlab#1{#1}\fi

\bibitem[{Bai et~al.(2022{\natexlab{a}})Bai, Jones, Ndousse, Askell, Chen,
  DasSarma, Drain, Fort, Ganguli, Henighan et~al.}]{bai2022training}
Yuntao Bai, Andy Jones, Kamal Ndousse, Amanda Askell, Anna Chen, Nova DasSarma,
  Dawn Drain, Stanislav Fort, Deep Ganguli, Tom Henighan, et~al.
  2022{\natexlab{a}}.
\newblock Training a helpful and harmless assistant with reinforcement learning
  from human feedback.
\newblock \emph{arXiv preprint arXiv:2204.05862}.

\bibitem[{Bai et~al.(2022{\natexlab{b}})Bai, Kadavath, Kundu, Askell, Kernion,
  Jones, Chen, Goldie, Mirhoseini, McKinnon et~al.}]{bai2022constitutional}
Yuntao Bai, Saurav Kadavath, Sandipan Kundu, Amanda Askell, Jackson Kernion,
  Andy Jones, Anna Chen, Anna Goldie, Azalia Mirhoseini, Cameron McKinnon,
  et~al. 2022{\natexlab{b}}.
\newblock Constitutional ai: Harmlessness from ai feedback.
\newblock \emph{arXiv preprint arXiv:2212.08073}.

\bibitem[{Bradley and Terry(1952)}]{bradley1952rank}
Ralph~Allan Bradley and Milton~E Terry. 1952.
\newblock Rank analysis of incomplete block designs: I. the method of paired
  comparisons.
\newblock \emph{Biometrika}, 39(3/4):324--345.

\bibitem[{Brown et~al.(2020)Brown, Mann, Ryder, Subbiah, Kaplan, Dhariwal,
  Neelakantan, Shyam, Sastry, Askell et~al.}]{brown2020language}
Tom Brown, Benjamin Mann, Nick Ryder, Melanie Subbiah, Jared~D Kaplan, Prafulla
  Dhariwal, Arvind Neelakantan, Pranav Shyam, Girish Sastry, Amanda Askell,
  et~al. 2020.
\newblock Language models are few-shot learners.
\newblock \emph{Advances in neural information processing systems},
  33:1877--1901.

\bibitem[{Chiang and Lee(2023)}]{chiang2023can}
Cheng-Han Chiang and Hung-yi Lee. 2023.
\newblock Can large language models be an alternative to human evaluations?
\newblock \emph{arXiv preprint arXiv:2305.01937}.

\bibitem[{Ding et~al.(2022)Ding, Qin, Liu, Chia, Joty, Li, and
  Bing}]{ding2022gpt}
Bosheng Ding, Chengwei Qin, Linlin Liu, Yew~Ken Chia, Shafiq Joty, Boyang Li,
  and Lidong Bing. 2022.
\newblock Is gpt-3 a good data annotator?
\newblock \emph{arXiv preprint arXiv:2212.10450}.

\bibitem[{Dubois et~al.(2023)Dubois, Li, Taori, Zhang, Gulrajani, Ba, Guestrin,
  Liang, and Hashimoto}]{dubois2023alpacafarm}
Yann Dubois, Xuechen Li, Rohan Taori, Tianyi Zhang, Ishaan Gulrajani, Jimmy Ba,
  Carlos Guestrin, Percy Liang, and Tatsunori~B Hashimoto. 2023.
\newblock Alpacafarm: A simulation framework for methods that learn from human
  feedback.
\newblock \emph{arXiv preprint arXiv:2305.14387}.

\bibitem[{Fernandes et~al.(2023)Fernandes, Madaan, Liu, Farinhas, Martins,
  Bertsch, de~Souza, Zhou, Wu, Neubig et~al.}]{fernandes2023bridging}
Patrick Fernandes, Aman Madaan, Emmy Liu, Ant{\'o}nio Farinhas, Pedro~Henrique
  Martins, Amanda Bertsch, Jos{\'e}~GC de~Souza, Shuyan Zhou, Tongshuang Wu,
  Graham Neubig, et~al. 2023.
\newblock Bridging the gap: A survey on integrating (human) feedback for
  natural language generation.
\newblock \emph{arXiv preprint arXiv:2305.00955}.

\bibitem[{Gilardi et~al.(2023)Gilardi, Alizadeh, and
  Kubli}]{gilardi2023chatgpt}
Fabrizio Gilardi, Meysam Alizadeh, and Ma{\"e}l Kubli. 2023.
\newblock Chatgpt outperforms crowd workers for text-annotation tasks.
\newblock \emph{Proceedings of the National Academy of Sciences},
  120(30):e2305016120.

\bibitem[{Guo et~al.(2023)Guo, Zhao, Tang, Zhao, and Wen}]{guo2023beyond}
Geyang Guo, Ranchi Zhao, Tianyi Tang, Wayne~Xin Zhao, and Ji-Rong Wen. 2023.
\newblock Beyond imitation: Leveraging fine-grained quality signals for
  alignment.
\newblock \emph{arXiv preprint arXiv:2311.04072}.

\bibitem[{Kullback and Leibler(1951)}]{kullback1951information}
Solomon Kullback and Richard~A Leibler. 1951.
\newblock On information and sufficiency.
\newblock \emph{The annals of mathematical statistics}, 22(1):79--86.

\bibitem[{Ouyang et~al.(2022)Ouyang, Wu, Jiang, Almeida, Wainwright, Mishkin,
  Zhang, Agarwal, Slama, Ray et~al.}]{ouyang2022training}
Long Ouyang, Jeffrey Wu, Xu~Jiang, Diogo Almeida, Carroll Wainwright, Pamela
  Mishkin, Chong Zhang, Sandhini Agarwal, Katarina Slama, Alex Ray, et~al.
  2022.
\newblock Training language models to follow instructions with human feedback.
\newblock \emph{Advances in Neural Information Processing Systems},
  35:27730--27744.

\bibitem[{Rafailov et~al.(2023)Rafailov, Sharma, Mitchell, Ermon, Manning, and
  Finn}]{rafailov2023direct}
Rafael Rafailov, Archit Sharma, Eric Mitchell, Stefano Ermon, Christopher~D
  Manning, and Chelsea Finn. 2023.
\newblock Direct preference optimization: Your language model is secretly a
  reward model.
\newblock \emph{arXiv preprint arXiv:2305.18290}.

\bibitem[{Ramamurthy et~al.(2022)Ramamurthy, Ammanabrolu, Brantley, Hessel,
  Sifa, Bauckhage, Hajishirzi, and Choi}]{ramamurthy2022reinforcement}
Rajkumar Ramamurthy, Prithviraj Ammanabrolu, Kiant{\'e} Brantley, Jack Hessel,
  Rafet Sifa, Christian Bauckhage, Hannaneh Hajishirzi, and Yejin Choi. 2022.
\newblock Is reinforcement learning (not) for natural language processing?:
  Benchmarks, baselines, and building blocks for natural language policy
  optimization.
\newblock \emph{arXiv preprint arXiv:2210.01241}.

\bibitem[{Schulman et~al.(2015)Schulman, Moritz, Levine, Jordan, and
  Abbeel}]{schulman2015high}
John Schulman, Philipp Moritz, Sergey Levine, Michael Jordan, and Pieter
  Abbeel. 2015.
\newblock High-dimensional continuous control using generalized advantage
  estimation.
\newblock \emph{arXiv preprint arXiv:1506.02438}.

\bibitem[{Schulman et~al.(2017)Schulman, Wolski, Dhariwal, Radford, and
  Klimov}]{schulman2017proximal}
John Schulman, Filip Wolski, Prafulla Dhariwal, Alec Radford, and Oleg Klimov.
  2017.
\newblock Proximal policy optimization algorithms.
\newblock \emph{arXiv preprint arXiv:1707.06347}.

\bibitem[{Touvron et~al.(2023)Touvron, Lavril, Izacard, Martinet, Lachaux,
  Lacroix, Rozi{\`e}re, Goyal, Hambro, Azhar et~al.}]{touvron2023llama}
Hugo Touvron, Thibaut Lavril, Gautier Izacard, Xavier Martinet, Marie-Anne
  Lachaux, Timoth{\'e}e Lacroix, Baptiste Rozi{\`e}re, Naman Goyal, Eric
  Hambro, Faisal Azhar, et~al. 2023.
\newblock Llama: Open and efficient foundation language models.
\newblock \emph{arXiv preprint arXiv:2302.13971}.

\bibitem[{Wang et~al.(2024)Wang, Zheng, Chen, Liu, Dou, Huang, Shen, Jin, Zhou,
  Shi et~al.}]{wang2024secrets}
Binghai Wang, Rui Zheng, Lu~Chen, Yan Liu, Shihan Dou, Caishuang Huang, Wei
  Shen, Senjie Jin, Enyu Zhou, Chenyu Shi, et~al. 2024.
\newblock Secrets of rlhf in large language models part ii: Reward modeling.
\newblock \emph{arXiv preprint arXiv:2401.06080}.

\bibitem[{Wang et~al.(2022)Wang, Kordi, Mishra, Liu, Smith, Khashabi, and
  Hajishirzi}]{wang2022self}
Yizhong Wang, Yeganeh Kordi, Swaroop Mishra, Alisa Liu, Noah~A Smith, Daniel
  Khashabi, and Hannaneh Hajishirzi. 2022.
\newblock Self-instruct: Aligning language models with self-generated
  instructions.
\newblock \emph{arXiv preprint arXiv:2212.10560}.

\bibitem[{Wu et~al.(2023)Wu, Hu, Shi, Dziri, Suhr, Ammanabrolu, Smith,
  Ostendorf, and Hajishirzi}]{wu2023fine}
Zeqiu Wu, Yushi Hu, Weijia Shi, Nouha Dziri, Alane Suhr, Prithviraj
  Ammanabrolu, Noah~A Smith, Mari Ostendorf, and Hannaneh Hajishirzi. 2023.
\newblock Fine-grained human feedback gives better rewards for language model
  training.
\newblock \emph{arXiv preprint arXiv:2306.01693}.

\bibitem[{Yuan et~al.(2023)Yuan, Yuan, Tan, Wang, Huang, and
  Huang}]{yuan2023rrhf}
Zheng Yuan, Hongyi Yuan, Chuanqi Tan, Wei Wang, Songfang Huang, and Fei Huang.
  2023.
\newblock Rrhf: Rank responses to align language models with human feedback
  without tears.
\newblock \emph{arXiv preprint arXiv:2304.05302}.

\bibitem[{Zheng et~al.(2023)Zheng, Chiang, Sheng, Zhuang, Wu, Zhuang, Lin, Li,
  Li, Xing et~al.}]{zheng2023judging}
Lianmin Zheng, Wei-Lin Chiang, Ying Sheng, Siyuan Zhuang, Zhanghao Wu, Yonghao
  Zhuang, Zi~Lin, Zhuohan Li, Dacheng Li, Eric Xing, et~al. 2023.
\newblock Judging llm-as-a-judge with mt-bench and chatbot arena.
\newblock \emph{arXiv preprint arXiv:2306.05685}.

\end{thebibliography}
\bibliographystyle{acl_natbib}

\newpage
\onecolumn
\appendix

\section{Appendix}
\label{sec:appendix}

\subsection{Prompt for Minimal Editing}
\label{sec:appendix1}
\begin{figure*}[h]
\centering 
\includegraphics[width=.99\linewidth]{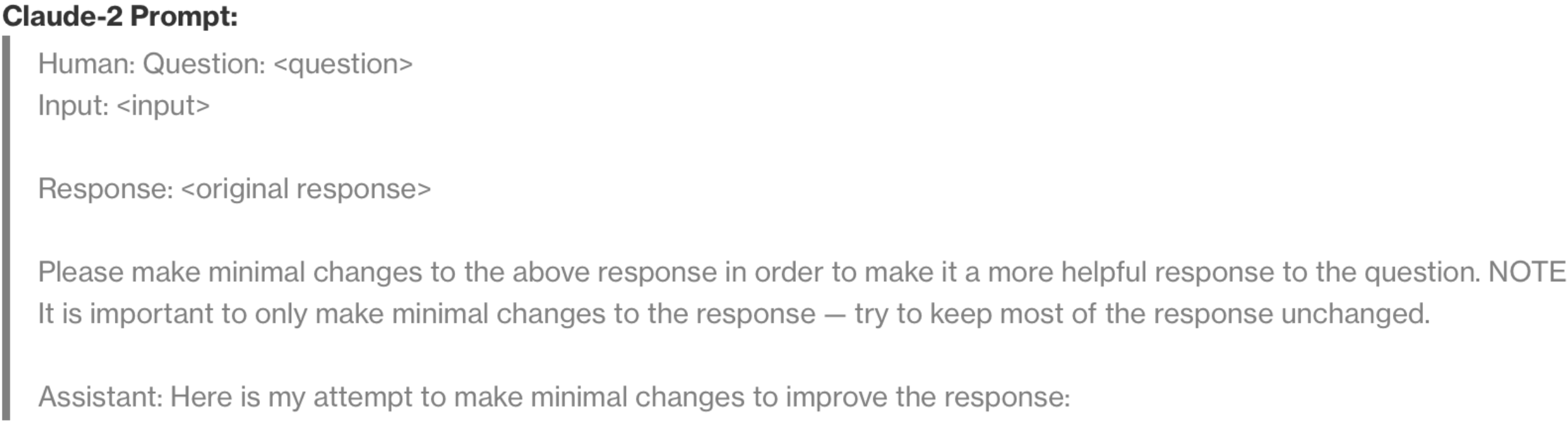} \\
  \caption{\small Prompt for \textit{Claude}
  }
  \label{fig:prompt}
\end{figure*}

Figure~\ref{fig:prompt} shows the prompt for \textit{Claude-2} to generate the fine-grained dataset by minimal editing. 

\subsection{More examples of minimal editing}
\label{sec:appendix2}
\begin{figure*}[h]
\centering 
\includegraphics[width=.9\linewidth]{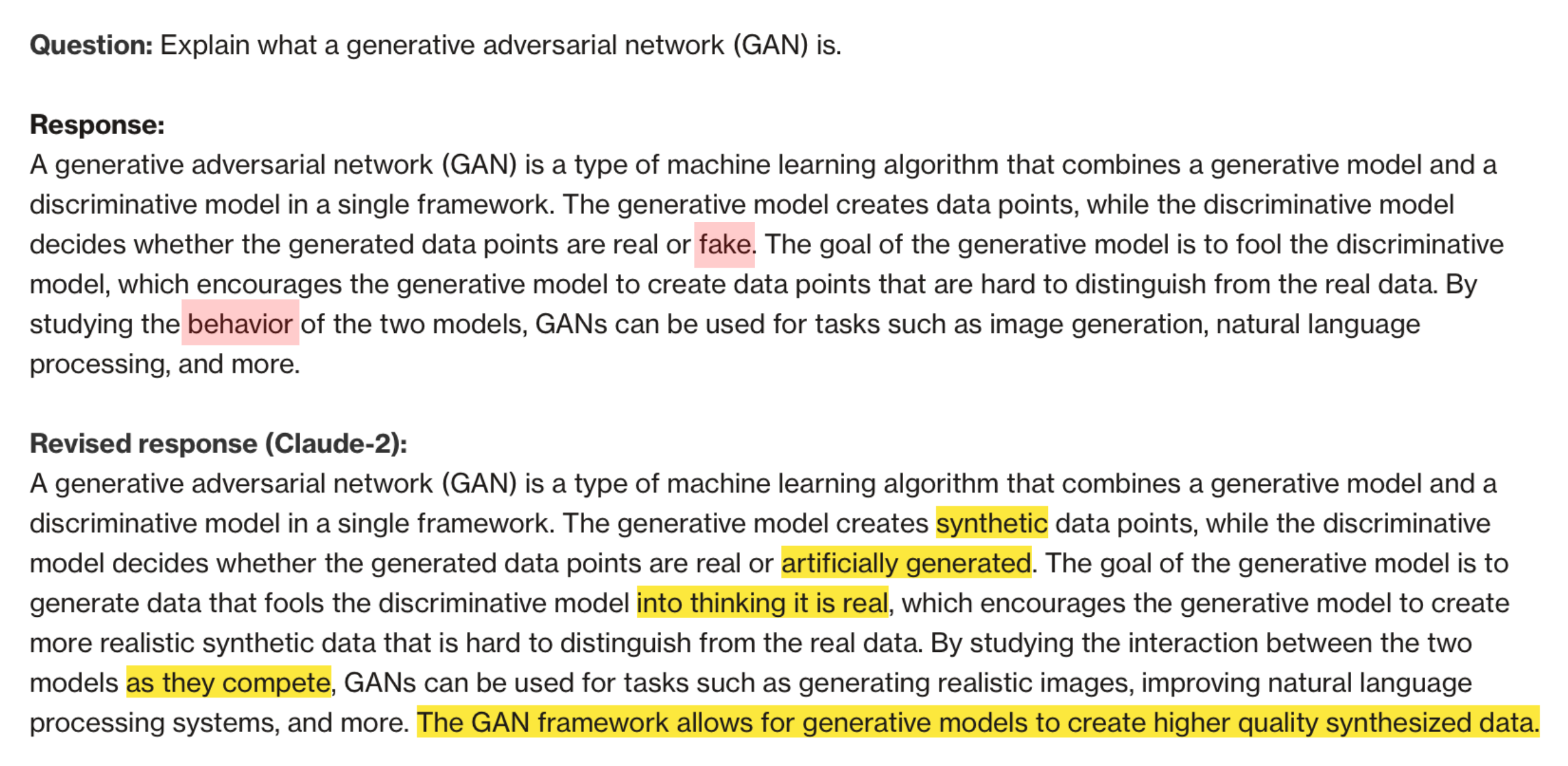} \\
  \caption{\small Example of fine-grained annotation via minimal editing: edit words may cause safety issues. }
  \label{fig:example1}
\end{figure*}

\begin{figure*}[h]
\centering 
\includegraphics[width=.9\linewidth]{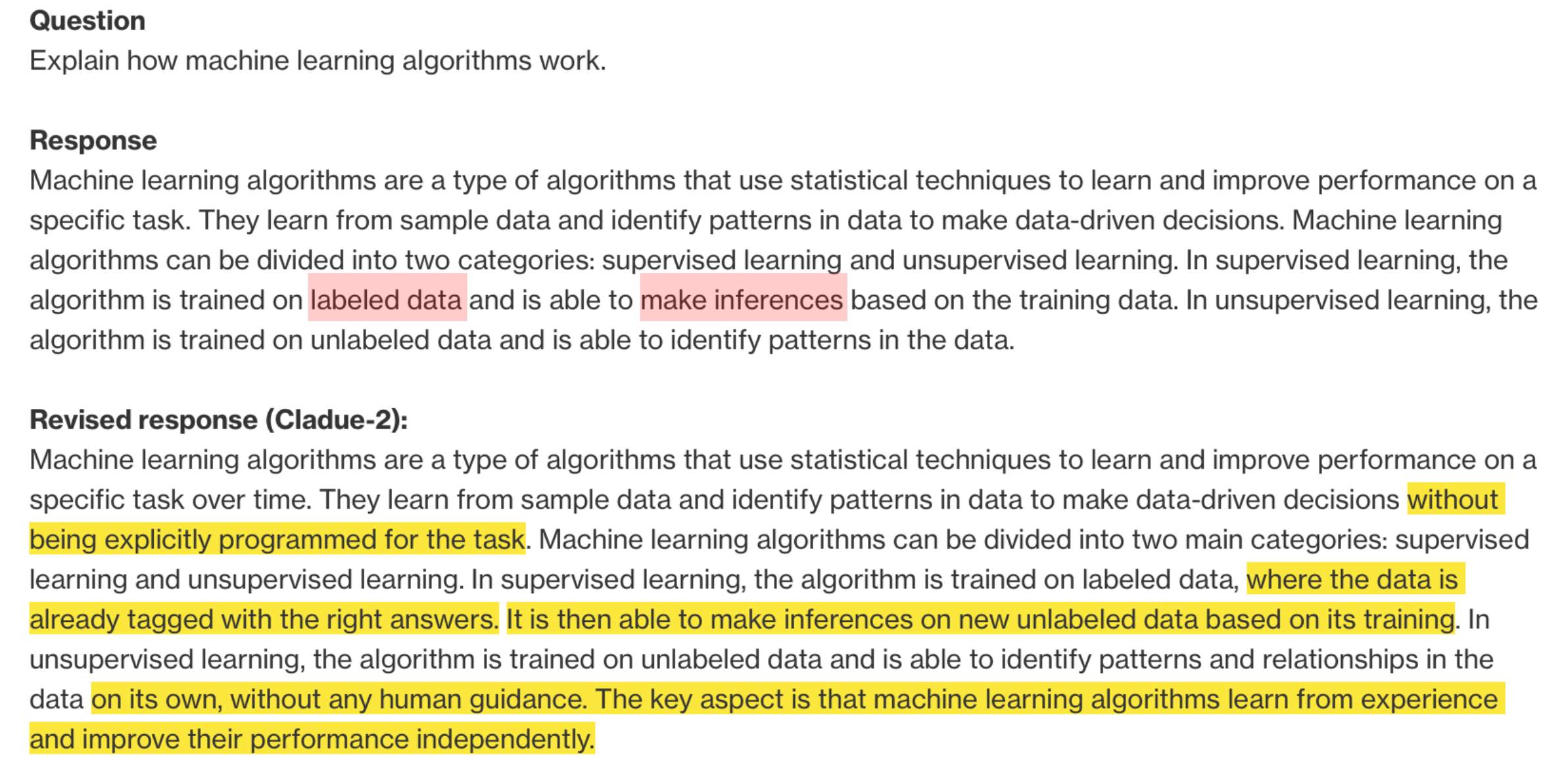} \\
  \caption{\small Example of fine-grained annotation via minimal editing: provide more explanation on academic words.  }
  \label{fig:example2}
\end{figure*}

\begin{figure*}[h]
\centering 
\includegraphics[width=.7\linewidth]{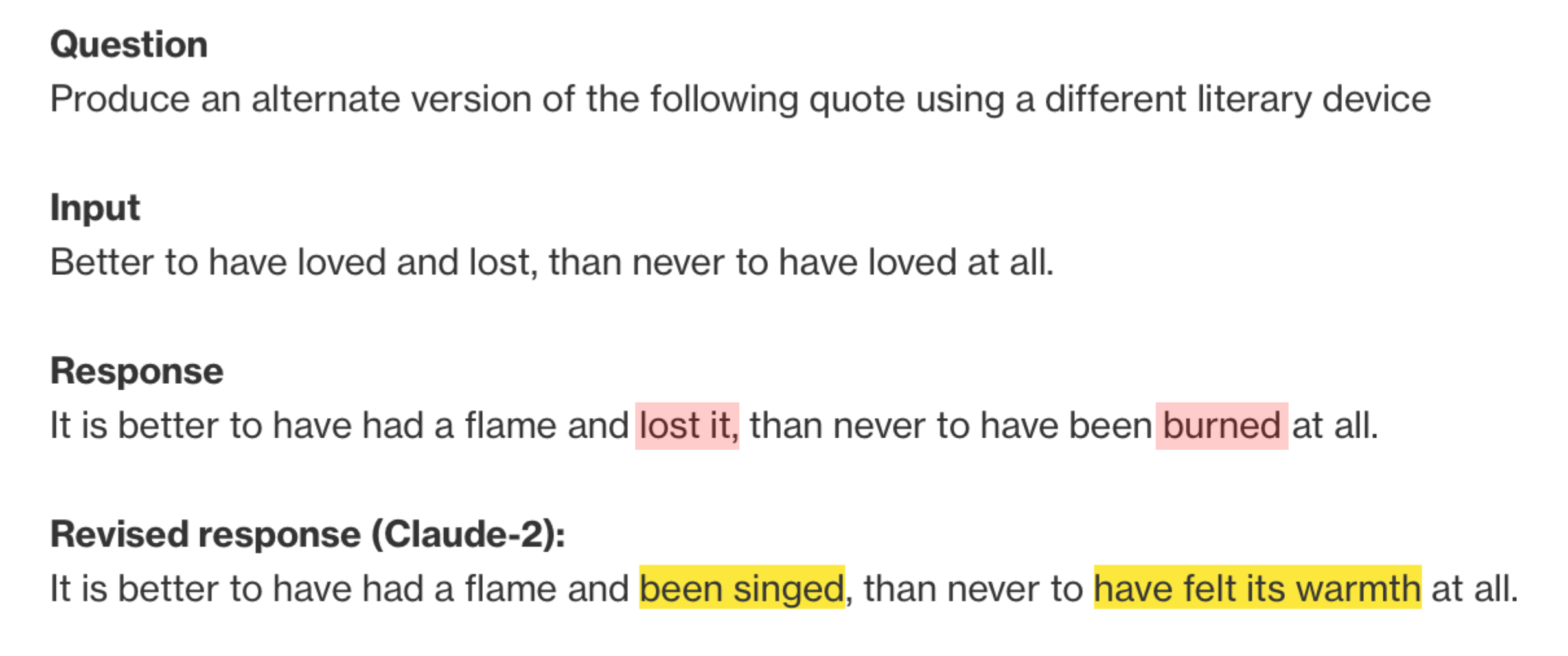} \\
  \caption{\small Example of fine-grained annotation via minimal editing: change the literary device that follows the instruction better. }
  \label{fig:example3}
\end{figure*}

In this section, we provide more examples of fine-grained annotation via minimal editing. As shown in Figure~\ref{fig:example1}, \ref{fig:example2} and \ref{fig:example3}, the annotators kept most of the response unchanged and only revised the red parts to the yellow parts. 

\end{document}